\pdfoutput=1
\def\isarxiv{1}
\documentclass[10pt,twocolumn,letterpaper]{article}
\usepackage[pagenumbers]{cvpr}
\usepackage{microtype}
\usepackage{graphicx}
\usepackage{booktabs,array}
\usepackage{amsmath,amssymb}
\usepackage{xspace}
\usepackage{pifont}
\usepackage{tikz}
\usetikzlibrary{positioning,arrows.meta,calc,fit,backgrounds,shapes.geometric,decorations.pathmorphing}
\definecolor{cvprblue}{rgb}{0.21,0.49,0.74}
\usepackage[pagebackref,breaklinks,colorlinks,allcolors=cvprblue]{hyperref}

\definecolor{geocol}{HTML}{3B6EA8}
\definecolor{appcol}{HTML}{C8742E}
\definecolor{bindcol}{HTML}{6A4C93}
\definecolor{inkgray}{HTML}{5A5A5A}
\definecolor{lightgray}{HTML}{EDEDED}

\newcommand{\suppref}[2]{\ifdefined\isarxiv #1\else #2\fi}
\colorlet{tokgray}{black!50}
\colorlet{tokmix}{appcol!60!bindcol}

\newcommand{\papertitle}{Ref-Renderer: Beyond Geometric Proxies for\\Appearance-Aware Video Worlds}
\newcommand{\methodname}{Ref-Renderer\xspace}

\def\confName{CVPR}
\def\confYear{2027}

\newcommand{\projecturl}{https://aawangas.github.io/ref-renderer/}

\newcommand{\arxivauthors}{%
  Anbang Wang$^{1,2}$ \quad
  Jiahao Lu$^{1}$ \quad
  Yongsen Mao$^{1}$ \quad
  Hongli Xu$^{2}$ \quad
  Weilong Yan$^{2}$\\[3pt]
  Chunyu Zou$^{2}$ \quad
  Siyu Hong$^{2}$ \quad
  Zhiyuan Zhang$^{2}$ \quad
  Jingwei Huang$^{2}$ \quad
  Yuan Liu$^{1}$\\[3pt]
  {\normalfont\normalsize $^{1}$The Hong Kong University of Science and Technology \quad $^{2}$Tencent}%
}

\title{\papertitle}
\author{\arxivauthors}
\newcommand{\teaserfigure}{%
\begingroup
\captionsetup{hypcap=false}
\centering
\includegraphics[width=\textwidth]{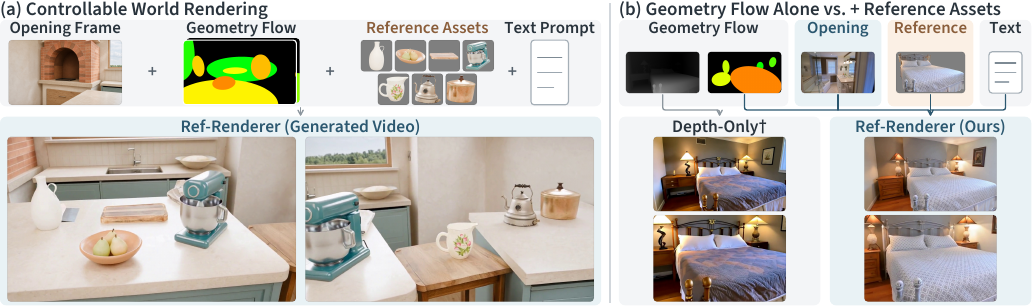}
\par
\vspace{-0.5em}
\captionof{figure}{\textbf{\methodname couples geometry flow with an independent asset bank to render specified appearances beyond the opening view.} \textbf{(a)} From a brick-oven opening frame with none of the seven kitchen assets visible, \methodname renders a coherent artisan bakery along a camera pan. \textbf{(b)} As an unseen bedroom enters from a bathroom opening view, a depth-only renderer ($^\dagger$Cosmos3-Nano~\cite{cosmos3}, given dense depth and text without references) preserves layout but hallucinates a dark-frame bed, whereas \methodname faithfully renders the referenced curved metal headboard and patterned bedding (zoomed below).}
\label{fig:teaser}
\vspace{0.25em}
\endgroup
}

\AtBeginDocument{%
  \crefname{appendix}{App.}{Apps.}%
  \Crefname{appendix}{Appendix}{Appendices}%
}

\newcommand{\maketitleappendix}{%
  \clearpage
  \twocolumn[
    \centering
    \Large
    \textbf{\thetitle}\\
    \vspace{0.5em}Appendix \\
    \vspace{1.0em}
  ]%
}

\begin{document}
\twocolumn[{\renewcommand\twocolumn[1][]{#1}\vspace{-1.2em}\maketitle\vspace{-1.4em}\teaserfigure\vspace{0.2em}}]
\begin{abstract}
Code world models, where coding agents or 3D programs construct a scene and explore it along a camera path, require specifying both how a world evolves across views and what each instance looks like. Existing renderers guide layout and camera motion with geometric proxies (such as depth maps, 3D boxes, or coarse meshes), assuming that the opening frame and text suffice to define appearance. This assumption breaks once exploration uncovers regions \textbf{outside the initial view}: newly emerging objects are absent from the opening frame, and neither coarse geometry nor text can specify exact visual identities, especially across multiple same-category instances. We present \methodname, which unifies time-varying structural proxies into a \emph{geometry flow} and couples it with an independent bank of cached appearance assets. Because the continuous geometry flow already governs cross-frame motion and occlusion, we avoid warping references into a per-frame video canvas that ties encoding to one trajectory, scales tokens with duration, and limits viewpoint adaptation. Instead, each asset claims its proxy via a single-frame anchor, mapping continuous coordinates into the target box in the backbone's native 3D RoPE space via unmodified joint self-attention. Decoupling appearance tokens from clip length enables modular asset editing and elastic latent readout across inference-time token budgets. On 266 real clips, \methodname matches a dense per-frame canvas in reference fidelity with $38\times$ fewer appearance tokens, while achieving accurate camera control from coarse geometry alone.
\ifdefined\isarxiv Project Page: \href{\projecturl}{\projecturl}.\fi
\end{abstract}

\section{Introduction}
\label{sec:intro}

Controllable video generation is shifting from single-shot motion guidance~\cite{cameractrl,das} toward \emph{code and agentic world models}, where coding agents or 3D programs construct a persistent scene and explore it along a camera path~\cite{genie,cwm,worlddirector} (\cref{fig:teaser}a). Across diverse upstream engines, time-varying structural proxies (such as depth maps, 3D boxes, or coarse meshes) can be unified into a \emph{geometry flow} that governs spatial layout, depth ordering, occlusion, and camera motion without specifying surface texture~\cite{blobgen,vace,cosmos3}. However, programmatic world exploration introduces a fundamental new challenge once the camera uncovers regions \textbf{outside the initial view}: \emph{geometric proxies dictate where objects move, but not what newly emerging instances look like}. Because objects outside the opening frustum are absent from the first frame, and global text cannot assign exact visual identities across multiple same-category instances, pure-geometry renderers are forced to hallucinate arbitrary appearances (\cref{fig:teaser}b).

Existing paradigms fall short of resolving this challenge. First, controllable 3D/4D world models~\cite{gen3c,lyra2,versecrafter,cwm} lift or track only the content visible in the opening image, providing no mechanism to specify identities for objects that enter later during exploration. Second, reference-to-video methods inject external subject identities~\cite{phantom,magref,alchemist,conceptmaster,skyreelsa2} and 2D trajectory cues~\cite{dreamvideo2,dreamvideoomni,tora2,alchemint}, but operate in 2D image or track space without a scene-wide 3D geometry flow for mutual occlusion and camera navigation. Third, a direct bridge is to project each object's 3D world position into per-frame 2D boxes along the camera path and warp its reference crop into every visible frame, forming a screen-space appearance video canvas~\cite{worlddirector}. However, per-frame warping ties appearance encoding to a single trajectory, scales sequence-concatenated appearance tokens linearly with video duration, and restricts 3D viewpoint and lighting adaptation by repeatedly supplying warped 2D views.

We overcome these limitations without per-frame warping by keeping appearance as an independent asset bank while letting the geometry flow govern temporal evolution. Our key insight is that \emph{spatiotemporal correspondence can be factorized into a single-frame identity handoff and geometry-driven propagation}: because the continuous geometry flow already tracks 3D motion and occlusion across frames, the renderer needs only a single-frame anchor to know which asset belongs to which geometric proxy.

\methodname realizes this factorization in a pretrained video diffusion transformer~\cite{minimaxh3}. Each reference is VAE-encoded once into a cached spatial latent. At a single visible anchor frame, we affinely map continuous normalized coordinates from the reference box into the target instance box within the backbone's native 3D rotary positional embedding (RoPE) space~\cite{rope}. The anchored assets, geometry flow, opening frame, and noisy target tokens interact through unmodified joint self-attention without auxiliary adapters, propagating the bound identity along the geometry flow. To train this renderer on real videos, we develop a task-tailored pipeline that constructs aligned depth-coded geometry flows, quality-filtered instance references, and single-frame anchors.

This decoupled formulation supports modular world authoring, including out-of-view object emergence and asset replacement without re-encoding unchanged assets, while keeping appearance tokens independent of video duration (reducing average full-budget appearance tokens from $32{,}256$ to $852$ per real clip). Operating in continuous RoPE space also enables \emph{elastic latent readout}, allowing a single model to consume spatially downsampled asset latents across inference-time token budgets. On 266 real clips, \methodname achieves reference-appearance fidelity comparable to a dense per-frame canvas baseline ($0.759$ vs.\ $0.765$ Ref-DINO-Tracked). Without explicit camera input, its coarse geometry flow also attains $1.173^\circ$ rotation error, competitive with camera-controlled generation.

Our contributions are:
\begin{itemize}
\item \textbf{Appearance-aware world rendering.} We decouple controllable world rendering into a continuous geometry flow and a trajectory-independent asset bank, resolving multi-instance identity and out-of-view object emergence~(\cref{sec:specification,sec:why}).
\item \textbf{Single-frame native-RoPE handoff.} We bind cached asset latents to geometric proxies at a single anchor frame in native 3D RoPE space without per-frame warping or auxiliary attention modules, enabling training-free elastic latent readout~(\cref{sec:binding,sec:joint}).
\item \textbf{Scalable curation and evaluation.} We build an automated pipeline that extracts aligned proxy flows, filtered assets, and anchors from real videos, matching dense per-frame canvas fidelity with $38\times$ fewer appearance tokens~(\cref{sec:data_pipeline,sec:experiments}).
\end{itemize}

\begin{figure*}[t]
\centering
\includegraphics[width=\textwidth]{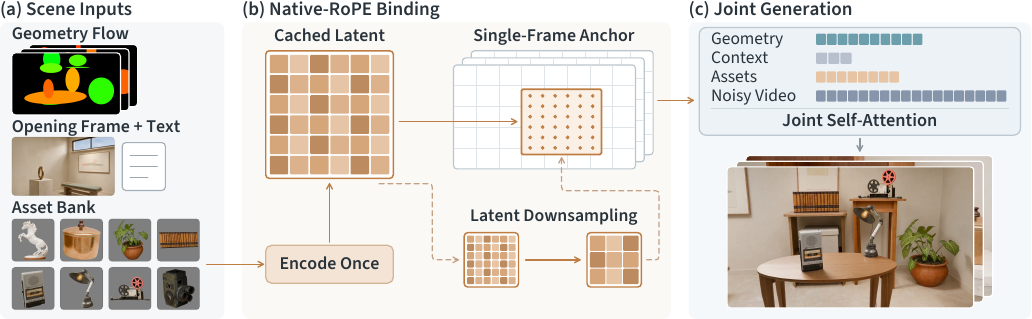}
\vspace{-0.6em}
\caption{\textbf{Overview of \methodname.} \textbf{(a)} Geometry flow specifies cross-frame scene evolution alongside an opening frame, text prompt, and instance asset bank. \textbf{(b)} Each asset is encoded once and bound to a single visible anchor by mapping its token coordinates into native 3D rotary positional embedding (RoPE) space; dashed paths indicate optional inference-time latent downsampling that preserves anchor location and extent. \textbf{(c)} Unmodified joint self-attention fuses geometry, context, bound assets, and noisy video tokens to render the clip (token grids and block sizes are schematic).}
\label{fig:method}
\vspace{-0.4em}
\end{figure*}

\section{Related Work}
\label{sec:related}

\paragraph{World models and structural video control.}
Generative world models synthesize interactive, explorable 3D environments~\cite{genie,wonderworld,aether}, with recent frameworks decoupling scene state and camera navigation from visual rendering via point-cloud caches and view retrieval~\cite{gen3c,lyra2,trajectorycrafter,worldstereo,captainsafari}, G-buffer or 3D tracking neural shaders~\cite{diffusionrenderer,das,uni3c,versecrafter}, agent-driven proxy videos~\cite{cwm}, or persistent dynamic memory~\cite{worlddirector}. Complementary structural controllers steer synthesis using camera poses~\cite{cameractrl}, 2D/3D layouts and entity trajectories~\cite{blobgen,vace,trajmaster3d,cinemaster,loosecontrolvideo,symphomotion}, or camera-aware RoPE adaptations~\cite{rerope,viewrope}. Although geometry flows govern 3D layout, motion, and occlusion, pure structural proxies cannot specify distinct appearances for multiple same-category instances or out-of-view objects emerging during exploration, whereas screen-space appearance warping~\cite{worlddirector} ties encoding to a single trajectory rather than maintaining reusable assets.

\paragraph{Reference-conditioned video generation.}
Subject-preserving and video-insertion models maintain reference identities across generated frames~\cite{phantom,magref,alchemist,conceptmaster,skyreelsa2,hunyuancustom,omniinsert}, while object-centric 3D proxy renderers~\cite{gorenderer} and trajectory- or shot-controlled methods associate entities with dense coordinate maps, spatial tracks, or temporal intervals~\cite{dreamvideo2,tora2,dreamvideoomni,alchemint,multishotmaster}. Existing DiT interfaces establish correspondence via dense per-frame coordinate concatenation~\cite{gorenderer}, auxiliary spatial or concept adapters~\cite{dreamvideo2,tora2,dreamvideoomni,conceptmaster}, shifted context-controller positional embeddings~\cite{ominicontrol,hunyuancustom,alchemint,poco}, or multi-frame box RoPE resampling that scales with annotated frames~\cite{multishotmaster}. Operating without a scene-wide 3D geometry flow also limits explicit depth ordering and camera exploration. In contrast, \methodname binds cached asset tokens to a geometric proxy at a single anchor frame in unmodified 3D RoPE space, delegating temporal propagation and occlusion to a scene-wide geometry flow without auxiliary adapters or per-frame coordinate maps.

\section{Ref-Renderer}
\label{sec:method}

\subsection{Overview}
\label{sec:specification}
Given a scene specification $\mathcal S$ exported by a coding agent, 3D authoring tool, or real-video parser, \methodname synthesizes a target video $V_{1:T}$ that follows the specified camera trajectory and scene dynamics while preserving instance identities~(\cref{fig:method}):
\begin{equation}
\mathcal S=\bigl(P,\;I_0,\;G_{1:T},\;\{(R_i,\tau_i,B_i)\}_{i=1}^{N}\bigr).
\label{eq:world}
\end{equation}
In \cref{eq:world}, $P$ is a scene text prompt, $I_0$ is the opening RGB frame, $G_{1:T}$ is a textureless \emph{geometry flow} specifying cross-frame structure and camera motion~(\cref{fig:method}a), and $\{(R_i,\tau_i,B_i)\}_{i=1}^{N}$ is an \emph{asset bank} pairing each reference image $R_i$ with its projected 2D bounding box $B_i$ at a single visible anchor frame $\tau_i\in\{1,\dots,T\}$.

Instead of warping $R_i$ into every frame along $G_{1:T}$, \methodname decouples appearance encoding from temporal propagation in two stages. First, each asset $R_i$ is encoded once into a trajectory-independent spatial latent, whose normalized token coordinates are affinely mapped into $B_i$ at frame $\tau_i$ within the backbone's native 3D RoPE space~(\cref{fig:method}b, \cref{sec:binding}). Second, the anchored asset tokens, geometry flow, opening frame, and noisy video tokens interact through unmodified joint self-attention, propagating each bound identity along $G_{1:T}$ across views and occlusions~(\cref{fig:method}c, \cref{sec:joint}).

\textbf{Geometry flow.}
Analogous to optical flow for 2D pixel motion, $G_{1:T}$ captures the time-varying screen-space projection of scene structure, conveying projected location, depth ordering, mutual occlusion, and frustum entry or exit. Although the interface accommodates diverse primitives (such as coarse meshes, 3D boxes, or depth maps), we instantiate $G_{1:T}$ as depth-coded elliptical proxies composited front-to-back on a black background: rendered 3D ellipsoids for authored worlds, and depth-associated 2D mask ellipses for real videos~(\cref{sec:geometry_data}). Because $G_{1:T}$ is devoid of surface texture, it dictates \emph{where} each proxy moves and occludes across $1:T$ while leaving \emph{what} each instance looks like to the asset bank. Since binding depends only on $(\tau_i,B_i)$, the same asset bank pairs directly with either sparse proxy flows or dense depth maps.

\textbf{Trajectory-independent asset bank.}
Each reference object $R_i$ is centered on a neutral square canvas with normalized asset bounding box $A_i$ and encoded once by the video VAE into a cached spatial latent $Z_i$. Encoded prior to camera projection, $Z_i$ remains invariant to camera trajectories, clip duration, and whether instance $i$ is visible in $I_0$, so replacing an object's design re-encodes only that single asset.

\subsection{Single-frame spatiotemporal binding}
\label{sec:binding}
To resolve which asset $R_i$ describes which trajectory in $G_{1:T}$ when multiple instances share a category or enter after $I_0$, we use a \emph{handoff design}: the anchor $(\tau_i,B_i)$ establishes identity-to-proxy correspondence at a single visible frame, and $G_{1:T}$ guides rendering across the remaining frames.

Specifically, for a reference token sampled at normalized coordinates $(u,v)$ on the asset canvas, we affinely map its relative position within $A_i=(a_x^0,a_y^0,a_x^1,a_y^1)$ into the target box $B_i=(b_x^0,b_y^0,b_x^1,b_y^1)$ at frame $\tau_i$:
\begin{align}
x&=b_x^0+\frac{u-a_x^0}{a_x^1-a_x^0}(b_x^1-b_x^0),\nonumber\\
y&=b_y^0+\frac{v-a_y^0}{a_y^1-a_y^0}(b_y^1-b_y^0).
\label{eq:map}
\end{align}
We express this anchored location directly in the backbone's native 3D rotary positional embedding (RoPE) coordinate space~\cite{rope,minimaxh3}:
\begin{equation}
\mathbf p_i(u,v)=\bigl(\rho_t(\tau_i),\rho_y(y),\rho_x(x)\bigr),
\label{eq:native}
\end{equation}
where $\rho_t$ maps $\tau_i$ to the temporal latent bin and $\rho_y,\rho_x$ map $(y,x)$ onto the continuous spatial grid. Native-RoPE binding injects $Z_i$ as a localized virtual observation at frame $\tau_i$; because $\rho_y(y)$ and $\rho_x(x)$ take continuous coordinates, binding requires neither screen-space rasterization nor discrete grid rounding.

\subsection{Joint conditioning and elastic latent readout}
\label{sec:joint}
Instantiated on the reference-conditioned pathway of MiniMax-H3~\cite{minimaxh3}, we patchify the cached asset latents, geometry flow, and opening frame into token sequences $C_i$, $C_G$, and $C_0$. Concatenated with noisy target video tokens $X_s$ at denoising step $s$, the transformer input is
\begin{equation}
\mathcal X_s=[C_G;C_0;C_1;\cdots;C_N;X_s].
\label{eq:sequence}
\end{equation}
Unmodified bidirectional joint self-attention operates across $\mathcal X_s$ in \cref{eq:sequence} alongside text conditioning, trained with a flow-matching loss on the target video. Unlike hard regional attention masks that discretize boxes, disrupt dense FlashAttention kernels, and block non-anchor frames from directly querying $C_i$, continuous 3D RoPE acts as a soft positional bias: it concentrates correspondence within $B_i$ at $\tau_i$ to disambiguate instances while preserving full cross-frame access to high-frequency asset details as $C_G$ evolves.

\textbf{Elastic latent readout (asset LoD).}
Because \cref{eq:map,eq:native} bind tokens via continuous normalized coordinates $(u,v)$ rather than fixed discrete slots, correspondence is independent of the spatial grid resolution of $Z_i$. At inference, a cached latent $Z_i$ can be spatially downsampled before patchification as an asset level-of-detail (LoD) control, recomputing $(u,v)$ under the same $A_i\to B_i$ map to adjust reference-token density around $B_i$ without retraining or altering anchor extent~(\cref{sec:exp_lod}).

\subsection{Why decouple: asset bank versus video canvas}
\label{sec:why}

\begin{figure}[t]
\centering
\includegraphics[width=\columnwidth]{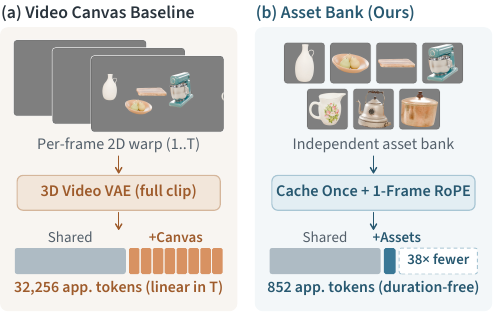}
\vspace{-0.35em}
\caption{\textbf{Video canvas versus decoupled asset bank.} \textbf{(a)} Canvas warps reference crops into projected 2D boxes across all frames $1\dots T$ and encodes the full clip ($32{,}256$ appearance tokens). \textbf{(b)} \methodname caches each asset once and binds it via single-frame RoPE ($852$ tokens on average, $38\times$ fewer and independent of duration; \cref{eq:conditioning_lengths}).}
\label{fig:why_decouple}
\vspace{-0.25em}
\end{figure}

To clarify the role of our representation, we contrast the decoupled asset bank against a dense screen-space appearance canvas under the shared H3 backbone and geometry flow $G_{1:T}$~(\cref{fig:why_decouple}). Given each instance's 3D trajectory, the canvas baseline projects its 3D location into per-frame 2D bounding boxes across all visible frames, warps the reference crop $R_i$ into each projected box, composites the warped crops into a full-frame conditioning video, encodes that video with the 3D VAE, and concatenates the resulting tokens alongside $C_G$.

\textbf{Duration-independent conditioning footprint.}
Unlike the canvas baseline whose conditioning token count grows linearly with video duration, our asset bank incurs a constant appearance-token cost regardless of clip length. Let $L_{\text{shared}}$ denote tokens shared by both variants (target video, geometry flow, opening frame, and text), $(H,W)$ the spatial token grid of a full-frame video latent over $T'$ latent frames, and $(h_i,w_i)$ the token grid for asset $i$:
\begin{equation}
\begin{aligned}
L_{\text{canvas}} &= L_{\text{shared}} + T'HW,\\
L_{\text{asset}} &= L_{\text{shared}} + \sum_{i=1}^{N} h_iw_i.
\end{aligned}
\label{eq:conditioning_lengths}
\end{equation}
Under quadratic self-attention, replacing $T'HW$ canvas tokens with $\sum_i h_iw_i$ asset tokens cuts sequence length and attention cost, with savings growing on longer clips~(\cref{sec:ablation}). Moreover, editing a camera path re-encodes only $G_{1:T}$ and updates $(\tau_i,B_i)$ while reusing all cached $\{Z_i\}$, and modifying an object's design re-encodes only that single asset.

\textbf{Generative prior and viewpoint adaptation.}
Single-frame anchoring preserves the generator's 3D appearance prior across viewpoints rather than locking synthesis to warped 2D crops. Whereas a per-frame canvas repeatedly injects a 2D-warped view at every frame, anchoring identity once at $\tau_i$ delegates novel-view synthesis to the backbone: guided structurally by textureless $G_{1:T}$, the model adapts shading, perspective deformation, and boundaries naturally to the evolving scene.

\section{Training Data Construction}
\label{sec:data_pipeline}
\begin{figure*}[t]
\centering
\includegraphics[width=\textwidth]{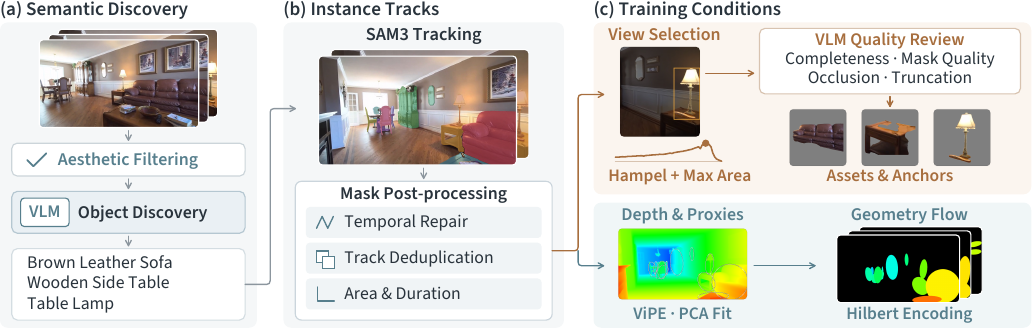}
\vspace{-0.6em}
\caption{\textbf{Training-data construction from real videos.} \textbf{(a)} After aesthetic filtering, a vision--language model (VLM) proposes object concepts from sampled frames. \textbf{(b)} SAM3 tracks instances across the clip, followed by temporal repair, deduplication, and support filtering. \textbf{(c)} Cleaned tracks branch into reference assets and anchors (via Hampel-filtered mask-area selection and VLM quality screening) and shared geometry flow (via ViPE depth, PCA ellipse fitting, and Hilbert depth-to-RGB rasterization), with the source clip retained as the supervision target.}
\label{fig:data_pipeline}
\vspace{-0.4em}
\end{figure*}

Training \methodname requires a scene-wide geometry flow $G_{1:T}$ together with instance references $\{R_i\}$ and single-frame anchors $\{(\tau_i,B_i)\}$ from the same scene. We construct these aligned conditions from real videos while retaining each source clip as the generation target. As shown in \cref{fig:data_pipeline}, our automated pipeline combines instance tracking with branches for asset selection and geometry flow.

\subsection{Video curation and instance tracking}
We curate videos from RealEstate10K~\cite{realestate10k} and Sekai~\cite{sekai} using aesthetic and image-quality filtering, partitioning clips by source video so training and held-out evaluation splits share no source videos or overlapping clips. A vision--language model (VLM) proposes object concepts from sampled frames to prompt SAM3~\cite{sam3} for instance segmentation and tracking. We repair temporal tracking artifacts, remove duplicate tracks, and filter instances with insufficient visibility or support, producing $42{,}482$ training clips ($166{,}052$ references) across $3{,}535$ source videos.

\subsection{Reference selection and quality filtering}
For each eligible track, we select a well-observed frame $\tau_i$ maximizing foreground mask area after Hampel filtering of isolated area spikes. The selected mask is cleaned, and a VLM screens the isolated crop for completeness and occlusion quality. We center the segmented foreground on a neutral gray canvas $R_i$ with asset box $A_i$ and pair it with its target bounding box $B_i$ at anchor frame $\tau_i$. Isolating the foreground on a neutral canvas removes surrounding scene context, while supervising the full clip forces the model to propagate identity from $\tau_i$ across changing viewpoints along the geometry track rather than copying background pixels. During training, we sample coarse scene-only captions ($30\%$) alongside detailed captions ($70\%$) so the model relies on visual assets over text for appearance.

\subsection{Geometry flow and paired examples}
\label{sec:geometry_data}
For tracked instances across the scene, we fit coarse 2D ellipses to valid masks using a robust PCA-based procedure and estimate depth with ViPE~\cite{vipe}. A Hilbert-style depth-to-RGB encoding and front-to-back depth-ordered rasterization turn these proxies into the shared geometry flow $G_{1:T}$, explicitly encoding mutual occlusion and frustum entry or exit as the camera moves. This condition captures evolving scene layout while leaving detailed instance appearance to the reference assets. Together, $G_{1:T}$, $\{R_i\}$, $\{(\tau_i,B_i)\}$, $I_0$, and $P$ form aligned conditions for reconstructing each source video clip.

\section{Experiments}
\label{sec:experiments}

\begin{table*}[t]
\caption{\textbf{Real-video comparison on 266 clips.} Best is in bold and second best is underlined. Coverage is in percent, rotation error in degrees, nMTE in percent, and VBench scores are shown as percentages. $^\dagger$These methods receive no reference images, so their Ref-DINO columns measure similarity to the evaluation targets. Baselines and metrics are described in \cref{sec:data}; all VBench dimensions are in \suppref{\cref{tab:supp_vbench}}{the supplement}.}
\label{tab:main}
\vspace{-0.4em}
\centering
\small
\setlength{\tabcolsep}{3pt}
\renewcommand{\arraystretch}{1.12}
\begin{tabular*}{\textwidth}{@{\extracolsep{\fill}}l cc c cc cccc@{}}
\toprule
 & \multicolumn{2}{c}{Ref-DINO} & & \multicolumn{2}{c}{Camera} & \multicolumn{4}{c}{VBench} \\
\cmidrule(lr){2-3}\cmidrule(lr){5-6}\cmidrule(l){7-10}
Method & Tracked $\uparrow$ & Fixed $\uparrow$ & Coverage $\uparrow$ & Rot.\ $\downarrow$ & nMTE $\downarrow$ & Subject $\uparrow$ & Backgr.\ $\uparrow$ & Flicker $\uparrow$ & Aesth.\ $\uparrow$ \\
\midrule
Lyra2$^\dagger$~\cite{lyra2} & 0.385 & 0.389 & 80.1 & \underline{0.801} & \underline{1.479} & 89.00 & 89.82 & \underline{95.04} & 52.16 \\
Cosmos3-Nano$^\dagger$~\cite{cosmos3} & 0.666 & \underline{0.657} & \textbf{99.0} & \textbf{0.271} & \textbf{0.536} & \underline{89.31} & \underline{91.96} & 94.54 & 50.48 \\
DreamVideo-Omni~\cite{dreamvideoomni} & 0.404 & 0.386 & 91.5 & 8.167 & 4.910 & 75.04 & 84.72 & 93.14 & 51.45 \\
H3 base~\cite{minimaxh3} & \underline{0.667} & 0.463 & 55.0 & 17.783 & 5.298 & 86.56 & 91.59 & 94.06 & \underline{53.39} \\
\midrule
\methodname & \textbf{0.759} & \textbf{0.701} & \underline{96.9} & 1.173 & 1.908 & \textbf{89.68} & \textbf{92.73} & \textbf{95.70} & \textbf{53.68} \\
\bottomrule
\end{tabular*}
\vspace{-0.2em}
\end{table*}

\begin{figure*}[t]
\centering
\includegraphics[width=\textwidth]{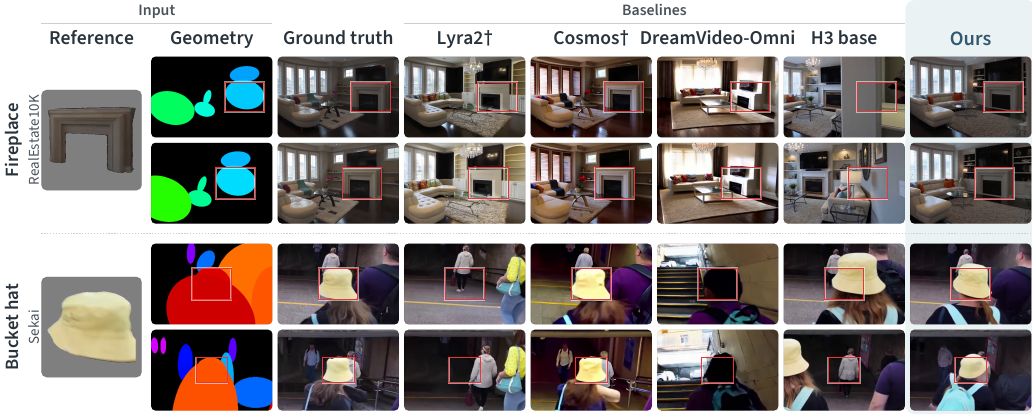}
\vspace{-0.5em}
\caption{\textbf{\methodname preserves reference identity as out-of-view objects enter the scene.} On two real clips, each row shows a zoom centered on the object's ground-truth location (red box), first as it enters the view and then after entry; the geometry flow is cropped identically. $^\dagger$Methods without reference-image input; Cosmos additionally receives dense source depth.}
\label{fig:baselines}
\end{figure*}

\begin{figure*}[t]
\centering
\includegraphics[width=\textwidth]{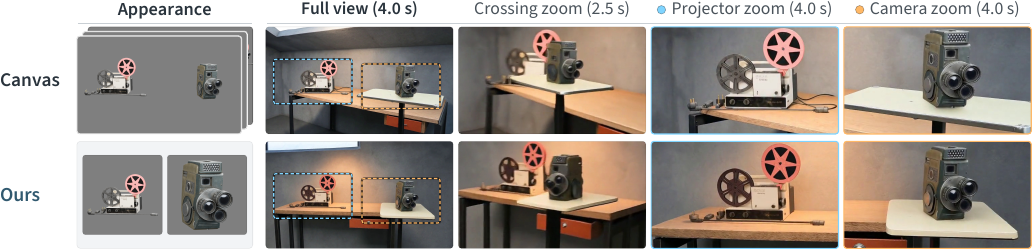}
\vspace{-0.5em}
\caption{\textbf{Per-frame canvas warping versus single-frame asset binding.} Both models receive the same opening frame (neither device in view), geometry flow, prompt, and two reference assets, supplied either as a warped per-frame video raster (Canvas, $32{,}256$ appearance tokens) or as independent gray-canvas assets (\methodname, $1{,}320$ tokens). As the camera crosses the doorway ($2.5\,\text{s}$) and settles in the editing bay ($4.0\,\text{s}$), Canvas exhibits ghosting under depth parallax, pasted 2D silhouettes, and warped support edges on the maple workbench and foreground camera stand, whereas \methodname preserves coherent 3D support geometry, warm directional shading, and grounded contact shadows.}
\label{fig:canvas_comparison}
\end{figure*}

\begin{table*}[t]
\caption{\textbf{Appearance-conditioning and token-budget ablations on 266 real clips.} Canvas and \methodname are trained separately under matched conditions; the three \methodname rows share a single checkpoint and vary the reference-token grid area only at inference. App.\ tokens report per-clip averages over the $266$ real clips; formatting and units follow \cref{tab:main}.}
\label{tab:ablation}
\vspace{-0.4em}
\centering
\small
\setlength{\tabcolsep}{3.2pt}
\renewcommand{\arraystretch}{1.12}
\begin{tabular*}{\textwidth}{@{\extracolsep{\fill}}l c cc c cc cccc@{}}
\toprule
 & & \multicolumn{2}{c}{Ref-DINO} & & \multicolumn{2}{c}{Camera} & \multicolumn{4}{c}{VBench} \\
\cmidrule(lr){3-4}\cmidrule(lr){6-7}\cmidrule(l){8-11}
Conditioning & App.\ tokens $\downarrow$ & Tracked $\uparrow$ & Fixed $\uparrow$ & Coverage $\uparrow$ & Rot.\ $\downarrow$ & nMTE $\downarrow$ & Subject $\uparrow$ & Backgr.\ $\uparrow$ & Flicker $\uparrow$ & Aesth.\ $\uparrow$ \\
\midrule
Canvas & 32{,}256 & 0.765 & 0.711 & 97.6 & 1.338 & 2.078 & 89.64 & 92.75 & 95.53 & 53.66 \\
\methodname\ 100\% & 852 & 0.759 & 0.701 & 96.9 & 1.173 & 1.908 & 89.68 & 92.73 & 95.70 & 53.68 \\
\methodname\ 50\% & 421 & 0.731 & 0.677 & 96.9 & 1.204 & 1.977 & 89.71 & 92.79 & 95.74 & 53.69 \\
\methodname\ 25\% & 213 & 0.736 & 0.677 & 97.1 & 1.232 & 2.056 & 89.75 & 92.73 & 95.74 & 53.73 \\
\bottomrule
\end{tabular*}
\vspace{-0.2em}
\end{table*}

\subsection{Experimental setup}
\label{sec:data}
\paragraph{Evaluation data.}
We evaluate \methodname on $266$ held-out real-video clips ($151$ RealEstate10K~\cite{realestate10k}, $115$ Sekai~\cite{sekai}; $127$ source videos), strictly disjoint from the $42{,}482$ training clips at both the clip and source-video levels, with $358$ reference entries annotated to enter after the opening frame. On real clips, each $R_i$ is cropped from source frame $\tau_i$ to test cross-view propagation as the camera uncovers new content; external synthetic and 3D-authored assets are evaluated in \cref{sec:exp_worlds} and \suppref{\cref{sec:supp_qual}}{the supplement}.

\paragraph{Baselines.}
We compare complementary control interfaces on the same clips with the full scene caption. Lyra2~\cite{lyra2} uses the opening image and estimated camera poses ($81$ frames, $832{\times}480$). Cosmos3-Nano~\cite{cosmos3} uses full-sequence depth estimated from the source video by ViPE~\cite{vipe}, without an opening frame or instance references ($121$ frames, $1280{\times}720$). DreamVideo-Omni~\cite{dreamvideoomni} uses reference images, boxes, and sparse point tracks, including background tracks for camera motion, without an opening frame ($49$ frames, $832{\times}480$). For H3 base~\cite{minimaxh3}, we replace our checkpoint with pretrained weights while retaining our opening-frame, geometry, asset-token, and RoPE layout, isolating task training on the same interface. H3 base and \methodname generate $107$ frames at $1344{\times}768$ ($W{\times}H$) from $672{\times}384$ geometry flows and $1024{\times}1024$ asset canvases.

Because the methods differ in frame count and resolution, we align controls and evaluation times by normalized video phase, matching each eligible source frame to the nearest output frame in phase. \Cref{tab:main} compares these control settings, whereas \cref{tab:ablation} isolates appearance representations under matched training. Lyra2 and Cosmos receive no reference images (marked $^\dagger$), and Cosmos additionally receives dense source-video depth.

\paragraph{Metrics.}
\emph{Appearance.} Ref-DINO measures how closely a generated instance matches its reference. To avoid scoring trivial reconstruction near $\tau_i$, we sample up to eight eligible source-visible frames strictly outside $\tau_i\pm 4$ frames and compute cosine similarity between foreground DINOv2~\cite{dinov2} features of the generated instance and its reference. Ref-DINO-Tracked locates the generated instance via a box-prompted SAM3~\cite{sam3} track initialized at the anchor box, omitting frames or instances without a qualifying mask. Ref-DINO-Fixed instead applies the ground-truth mask directly to the generated frame, penalizing spatial misplacement without relying on output tracking. Coverage is the fraction of eligible frames where the tracked mask meets the minimum size and overlaps the ground-truth box (\suppref{thresholds in \cref{sec:supp_eval}}{thresholds in supplement}). For Lyra2 and Cosmos, which take no reference images, Ref-DINO measures similarity to the ground-truth target crop.

\emph{Camera.} ViPE estimates trajectories from both source and generated videos. Rotation error uses first-frame-relative orientations; translation nMTE is the mean camera-center error after $\mathrm{Sim}(3)$ alignment, normalized by source path length. Both are evaluated at $81$ uniform phases.

\emph{Video quality.} From VBench~\cite{vbench}, \cref{tab:main} reports subject consistency, background consistency, temporal flickering, and aesthetic quality; all seven appear in \suppref{\cref{tab:supp_vbench}}{the supplement}.

\subsection{Main results}
\label{sec:exp_real}
\paragraph{Appearance grounding.}
\methodname achieves $0.759$ Ref-DINO-Tracked, $0.701$ Ref-DINO-Fixed, and $96.9\%$ coverage across the $266$ real-video clips~(\cref{tab:main}). DreamVideo-Omni, which relies on sparse 2D trajectories without an opening RGB frame, obtains $0.404$ Ref-DINO-Tracked and $91.5\%$ coverage. H3 base reaches $0.667$ Ref-DINO-Tracked at only $55.0\%$ tracking coverage, dropping to $0.463$ on Ref-DINO-Fixed because Tracked scores only frames with a qualifying mask and omits missing or misplaced predictions; under a stricter mask IoU $\ge 0.3$ criterion, \methodname retains $91.3\%$ coverage versus $28.7\%$ for H3 base (\suppref{see \cref{sec:supp_eval}}{see supplement}). Lyra2 and Cosmos, which take no reference images, attain $0.385$ and $0.666$ target similarity, respectively, with Cosmos aided by dense depth and detailed captions. \Cref{fig:baselines} shows two clips where the target enters after $I_0$: \methodname renders the referenced fireplace and yellow bucket hat in the expected regions with consistent scene lighting, whereas H3 base misses the fireplace, Cosmos hallucinates mismatched surface details despite dense depth, and Lyra2 and DreamVideo-Omni fail to reproduce the hat.

\paragraph{Camera control and video quality.}
\methodname achieves $1.173^\circ$ rotation error and $1.908\%$ translation nMTE, outperforming Canvas, DreamVideo-Omni, and H3 base. Lyra2 ($0.801^\circ$, $1.479\%$) receives explicit target camera poses and Cosmos ($0.271^\circ$, $0.536\%$) receives dense source-video depth, whereas our model infers camera motion from coarse elliptical proxies alone. Dense depth likewise accounts for Cosmos's $99.0\%$ coverage by fixing per-pixel scene layout. Across VBench metrics, \methodname achieves the highest subject consistency ($89.68$), background consistency ($92.73$), temporal flickering ($95.70$), and aesthetic quality ($53.68$), indicating that single-frame asset injection preserves temporal coherence across camera motion.

\subsection{Ablation and efficiency analysis}
\label{sec:ablation}
\paragraph{Appearance representation.}
We train \methodname and a dense Canvas baseline under matched conditions on the same H3 backbone, dataset, and schedule (\suppref{\cref{sec:supp_impl}}{supplement}), supplying identical evaluation inputs. Canvas warps references into projected boxes at every visible frame and encodes the resulting video ($32{,}256$ conditioning tokens per clip), whereas \methodname uses single-frame anchored asset tokens.

Full-budget \methodname reaches $0.759$ Ref-DINO-Tracked against $0.765$ for Canvas ($\Delta = -0.006$), closely matching tracked fidelity while using $852$ versus $32{,}256$ appearance tokens per clip~(\cref{tab:ablation}). Under H3's sequence-concatenated conditioning (unlike channel-concatenated canvases~\cite{worlddirector}), reducing appearance tokens by $38\times$ ($32{,}256 \to 852$) shortens visual token sequence length from $73.6\text{k}$ to $42.2\text{k}$ tokens ($1.74\times$) and yields an ${\approx}2\times$ generation speedup on $107$-frame clips, with savings growing on longer clips as Canvas scales with duration. Ref-DINO-Fixed is $0.010$ lower ($0.701$ vs.\ $0.711$), reflecting a natural trade-off of single-frame anchoring: per-frame pasting repeatedly injects the exact reference crop, favoring pixel-aligned DINO similarity, whereas our renderer avoids parallax ghosting and adapts 3D viewpoint, shading, and support boundaries to the scene~(\cref{fig:canvas_comparison}).

\paragraph{Elastic latent readout across token budgets.}
\label{sec:exp_lod}
Using a single checkpoint, we spatially downsample cached asset latents to retain $100\%$, $50\%$, or $25\%$ of the reference-token grid area, recomputing token centers under the same affine anchor map. This lowers average appearance tokens per clip from $852$ to $421$ and $213$, providing training-free headroom in multi-asset or memory-constrained scenes as total asset tokens scale with $N$. Quality degrades gracefully and saturates between $50\%$ and $25\%$: Ref-DINO-Tracked moves to $0.731$ and $0.736$ ($0.677$ Ref-DINO-Fixed for both), while coverage ($96.9\%$--$97.1\%$), camera accuracy, and VBench scores remain virtually unchanged, trading only fine high-frequency surface detail for a $4\times$ smaller asset footprint (\suppref{visual comparison in \cref{fig:budget}}{visual comparison in the supplement}).

\begin{figure}[t]
\centering
\includegraphics[width=\columnwidth]{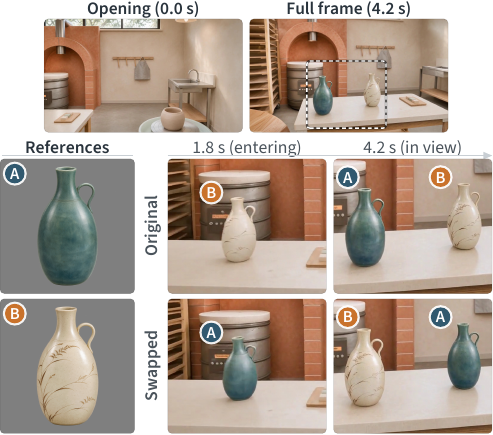}
\vspace{-0.5em}
\caption{\textbf{Reference swap under fixed geometry.} Holding the opening frame ($0.0\,\text{s}$, neither pitcher in view), geometry flow, prompt, and seed fixed, we swap only the smoke-celadon ($A$) and ivory grass-motif ($B$) references assigned to two ceramic pitchers on the stone bench. Crops (dashed box at $4.2\,\text{s}$) show entry ($1.8\,\text{s}$) and clean identity exchange ($4.2\,\text{s}$) at the same 3D positions.}
\label{fig:swap}
\vspace{-0.4em}
\end{figure}

\subsection{Reference swap and authored worlds}
\label{sec:exp_worlds}
To verify multi-instance disambiguation under single-frame RoPE anchoring, we hold the opening frame, geometry flow, text prompt, and noise seed fixed while exchanging only the two external gray-canvas references $(A, B)$ assigned to two same-category proxies~(\cref{fig:swap}). Neither ceramic pitcher is visible in the opening frame; as the camera brings both into view, their appearances swap cleanly while 3D positions, scales, and bench contacts stay locked to the geometry flow (unanchored background shelving varies slightly under changed conditioning). \suppref{\Cref{sec:supp_qual}}{The supplement} provides additional multi-instance swaps across vases, clocks, and lamps, plus full-clip real and authored 3D world renderings with external assets under occlusion and large viewpoint shifts.

\section{Conclusion}
\label{sec:conclusion}
We have presented \methodname, a controllable world renderer that decouples evolving scene structure from instance appearance by pairing a continuous geometry flow with a trajectory-independent bank of cached appearance assets. Binding each asset to its proxy at a single visible anchor frame in native 3D RoPE space avoids dense per-frame canvas warping, operates via unmodified joint self-attention, and supports training-free elastic latent readout across inference-time token budgets. Across $266$ held-out real-world clips, multi-instance reference swaps, and authored 3D worlds, single-frame anchoring matches dense per-frame canvas fidelity with $38\times$ fewer appearance tokens and ${\approx}2\times$ faster $107$-frame generation while preserving accurate camera trajectories. Because aggressive latent downsampling softens fine surface textures at low token budgets and global camera motion is inferred implicitly from sparse proxy flows, coupling explicit camera cues with adaptive multi-scale asset tokens is a promising direction for future work.

{\small
\bibliographystyle{ieeenat_fullname}
\bibliography{references}
}
\maketitleappendix
\appendix
\setcounter{figure}{0}
\setcounter{table}{0}
\setcounter{equation}{0}
\renewcommand{\thefigure}{S\arabic{figure}}
\renewcommand{\thetable}{S\arabic{table}}
\renewcommand{\theequation}{S\arabic{equation}}
\renewcommand{\theHfigure}{S\arabic{figure}}
\renewcommand{\theHtable}{S\arabic{table}}
\renewcommand{\theHequation}{S\arabic{equation}}
\section{Implementation and Training Details}
\label{sec:supp_impl}
\paragraph{Temporal binning and spatial RoPE mapping.}
\methodname maps continuous anchor times onto the backbone's temporal latent bins, whose widths follow the repeating frame pattern $(1,4,4,4,4)$ across the $0$-indexed full clip $t\in\{0,\dots,T\}$ ($I_0$ at $t{=}0$, $T{=}106$ for $107$-frame clips). Let $e_0=0$ and $e_{k+1}=e_k+d_k$, where $d_k$ follows this pattern; a frame position $\tau_i$ falls into the unique bin satisfying $e_k\leq\tau_i<e_{k+1}$, and the reference tokens adopt that bin's temporal RoPE coordinate while retaining fractional $\tau_i$ until lookup to avoid frame-rate rounding drift. Spatially, each asset crop on the $1024{\times}1024$ canvas is aligned to $2{\times}2$ latent patch boundaries: normalized horizontal edges $a_x^0,a_x^1$ on a latent grid of width $W_s$ snap to $2\lfloor a_x^0W_s/2\rfloor$ and $2\lceil a_x^1W_s/2\rceil$ (and analogously vertically), with area interpolation preserving patch divisibility under elastic latent readout (asset LoD). Patch-center coordinates $(u,v)$ are computed on the original asset canvas and passed through the asset-to-target affine map so the small alignment margin is preserved rather than stretched into the object box. For a target token grid of size $H_g{\times}W_g$ with native spatial RoPE origins $o_y,o_x$ and increments $\Delta_y,\Delta_x$, the continuous coordinate maps in the main paper take the form
\begin{align}
\rho_y(y)&=o_y+(yH_g-\tfrac12)\Delta_y,\label{eq:supp_rope_y}\\
\rho_x(x)&=o_x+(xW_g-\tfrac12)\Delta_x,\label{eq:supp_rope_x}
\end{align}
preserving aspect-dependent coordinate scale across resolutions.

\paragraph{Matched training and token accounting.}
\methodname and the dense Canvas baseline are trained from the same pretrained H3 backbone on the same $42{,}482$ clips for $3{,}000$ optimization steps under identical opening frames, captions (sampling full scene captions and coarse scene-only captions with probabilities $0.7$ and $0.3$), and geometry flows: Canvas warps each reference into its projected box at every visible frame and VAE-encodes the resulting $1344{\times}768$ video ($32{,}256$ conditioning tokens per clip), whereas \methodname encodes each asset once and binds its tokens at a single anchor $\tau_i$. Both models train rank-$64$ LoRA adapters ($\alpha{=}64$, zero dropout) on attention query, key, value, and output projections plus a linear video residual head using AdamW ($\beta{=}(0.9,0.999)$, zero weight decay, learning rate $10^{-5}$, $100$ linear warmup steps then constant schedule, gradient clip $1.0$), hybrid-sharded FSDP2 in bf16, and global batch size $24$ ($107$ frames at $1344{\times}768$) under a CFG-aware flow-matching objective (guidance scale $5$). At inference, both models use $50$ steps, flow shift $12$, and seed $42$, while DreamVideo-Omni is evaluated with its public SFT checkpoint and temporal indices aligned to $0$ and the latent clip length. Averaged over the $266$ real clips, \methodname uses $852$, $421$, and $213$ appearance tokens at the $100\%$, $50\%$, and $25\%$ asset LoD budgets versus $32{,}256$ for Canvas. Combined with $32{,}256$ noisy target tokens and $9{,}072$ shared visual conditioning tokens (opening frame and geometry flow), visual token sequence length drops from $73.6\text{k}$ ($73{,}584$, Canvas) to $42.2\text{k}$ tokens ($42{,}180$, full-budget \methodname), yielding an ${\approx}2\times$ generation speedup on $107$-frame clips while lower budgets free multi-asset headroom.

\section{Evaluation Protocol and Extended Results}
\label{sec:supp_eval}
\begin{table*}[t]
\caption{\textbf{All seven VBench dimensions on 266 real-video clips.} Scores are reported as percentages; Dynamic measures motion magnitude rather than synthesis quality, and $^\dagger$denotes methods without reference-image input.}
\label{tab:supp_vbench}
\vspace{-0.4em}
\centering
\small
\setlength{\tabcolsep}{4pt}
\begin{tabular*}{\textwidth}{@{\extracolsep{\fill}}l ccccccc@{}}
\toprule
Method & Subject $\uparrow$ & Background $\uparrow$ & Smoothness $\uparrow$ & Flicker $\uparrow$ & Dynamic & Aesthetic $\uparrow$ & Imaging $\uparrow$ \\
\midrule
Ground truth & 89.77 & 93.58 & 98.10 & 95.13 & 99.25 & 53.09 & 74.61 \\
\midrule
Lyra2$^\dagger$ & 89.00 & 89.82 & 98.65 & 95.04 & 98.87 & 52.16 & 72.63 \\
Cosmos3-Nano$^\dagger$ & 89.31 & 91.96 & 97.33 & 94.54 & 99.62 & 50.48 & 75.32 \\
DreamVideo-Omni & 75.04 & 84.72 & 97.52 & 93.14 & 99.62 & 51.45 & 68.14 \\
H3 base & 86.56 & 91.59 & 98.44 & 94.06 & 99.62 & 53.39 & 72.02 \\
Canvas & 89.64 & 92.75 & 98.46 & 95.53 & 98.87 & 53.66 & 73.35 \\
\methodname\ 100\% & 89.68 & 92.73 & 98.53 & 95.70 & 99.25 & 53.68 & 72.99 \\
\methodname\ 50\% & 89.71 & 92.79 & 98.55 & 95.74 & 98.87 & 53.69 & 72.94 \\
\methodname\ 25\% & 89.75 & 92.73 & 98.54 & 95.74 & 98.87 & 53.73 & 73.06 \\
\bottomrule
\end{tabular*}
\vspace{-0.2em}
\end{table*}

\paragraph{Ref-DINO and tracking protocol.}
For each evaluated frame, the foreground mask bounding box is expanded by $1.05\times$ into a square crop, outside-mask pixels are set to neutral gray ($128$), and the crop is resized to $224{\times}224$; DINOv2-giant with registers extracts $16{\times}16$ patch tokens that are pooled by foreground-area weights and $\ell_2$-normalized. Source frame $t\in\{0,\dots,106\}$ is matched by phase to generated frame $\lfloor t/106\cdot(n-1)+0.5\rfloor$ for an $n$-frame output, and a frame is eligible when its ground-truth instance mask covers at least $256/(1344\cdot 768)$ of the frame and lies strictly more than $4$ frames from $\tau_i$ on the $107$-frame timeline (Ref-DINO samples up to eight evenly spaced eligible frames; coverage evaluates all eligible frames). Per-frame cosine similarity to the reference embedding is averaged across evaluated frames, then across duplicate entries of the same physical instance, and finally across clips.

\paragraph{Coverage threshold sensitivity.}
SAM3 is prompted solely with the annotated anchor bounding box (without text prompts, preventing cross-instance leakage among same-category objects) on both source and generated videos. By default, a tracked mask is accepted when it meets the minimum area criterion and its bounding-box intersection with the ground-truth box is at least $5\%$ of the smaller box. Across stricter criteria ($\text{overlap}\ge 0.05$, $\text{overlap}\ge 0.50$, $\text{box IoU}\ge 0.30$, $\text{box IoU}\ge 0.50$, and $\text{mask IoU}\ge 0.30$), method rankings remain unchanged: on the $266$ real clips, full-budget \methodname achieves $96.9\%$, $95.9\%$, $93.4\%$, $89.3\%$, and $91.3\%$ coverage, respectively, whereas H3 base drops from $55.0\%$ to $43.4\%$, $37.2\%$, $24.3\%$, and $28.7\%$.

\paragraph{Full VBench evaluation.}
\Cref{tab:supp_vbench} reports all seven VBench dimensions across the $266$ real-video clips, extending the four main-paper metrics. Dynamic degree measures motion magnitude rather than synthesis quality, so values near ground truth ($99.25\%$) reflect matched scene dynamics.

\begin{figure}[t]
\centering
\includegraphics[width=\columnwidth]{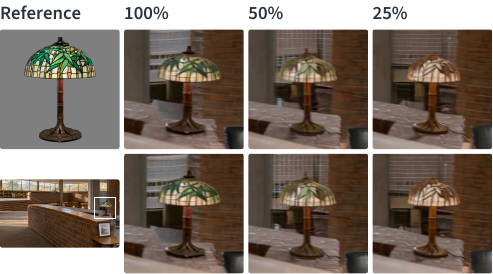}
\vspace{-0.5em}
\caption{\textbf{Elastic latent readout (asset LoD) across inference-time token budgets.} Corresponding frames generated from a single checkpoint using $100\%$, $50\%$, and $25\%$ of the full reference-token grid area. The left column shows the reference asset and scene context (white box marks the zoom region). Spatial placement remains stable across LoD budgets, while lampshade saturation softens at lower token densities.}
\label{fig:budget}
\vspace{-0.4em}
\end{figure}

\section{Additional Qualitative Results}
\label{sec:supp_qual}
\begin{figure*}[p]
\centering
\includegraphics[width=\textwidth]{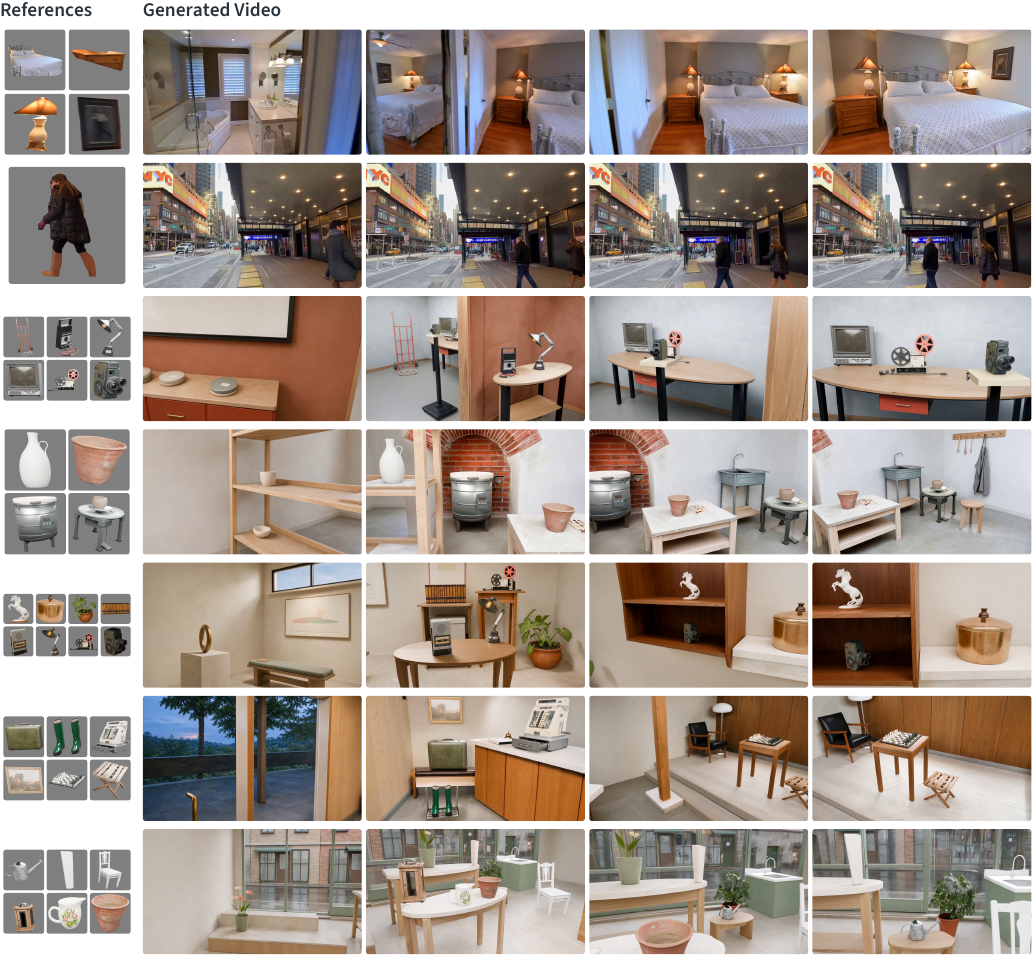}
\vspace{-0.4em}
\caption{\textbf{Qualitative world rendering across real and authored scenes.} Seven videos generated by \methodname at the full reference budget, showing input assets and time-ordered frames. The top two rows use held-out real-video clips (four and one reference assets). The bottom five rows use authored 3D scenes with six, four, eight, six, and six reference assets: an analog film studio, a ceramic studio, a collector's gallery, a twilight inn reception and chess lounge, and a rainy florist shop.}
\label{fig:world_examples}
\end{figure*}

\begin{figure*}[p]
\centering
\includegraphics[width=\textwidth]{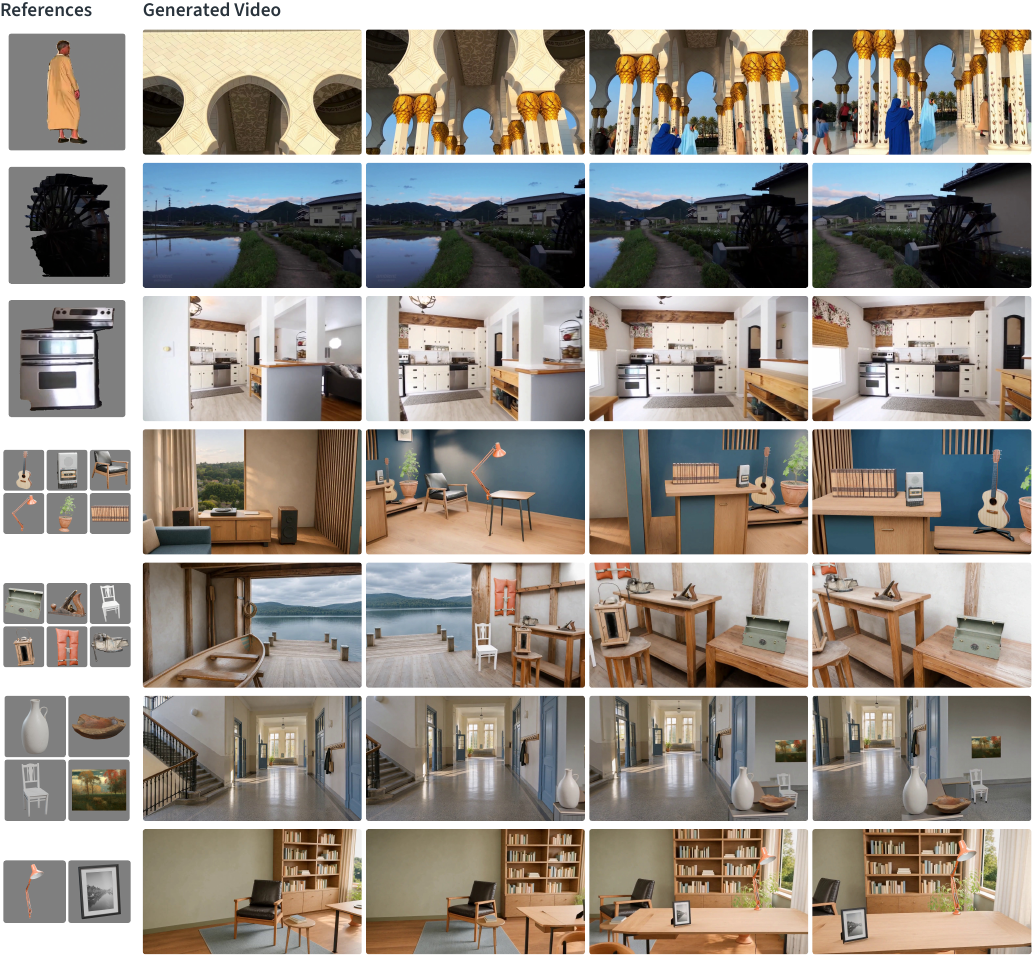}
\vspace{-0.4em}
\caption{\textbf{Additional world renderings across real and authored scenes.} Full-budget \methodname outputs showing reference assets on neutral-gray canvases and four time-ordered frames as off-screen targets enter under camera rotation and translation. The top three rows use held-out Sekai and RealEstate10K clips (a mosque colonnade, a waterside mill, and a sunlit kitchen); the bottom four rows use multi-asset authored 3D scenes with six, six, four, and two reference assets (a listening room, a lakeside boathouse workshop, a sculpture gallery, and a sunlit study).}
\label{fig:worlds_more}
\end{figure*}

\begin{figure*}[p]
\centering
\includegraphics[width=\textwidth]{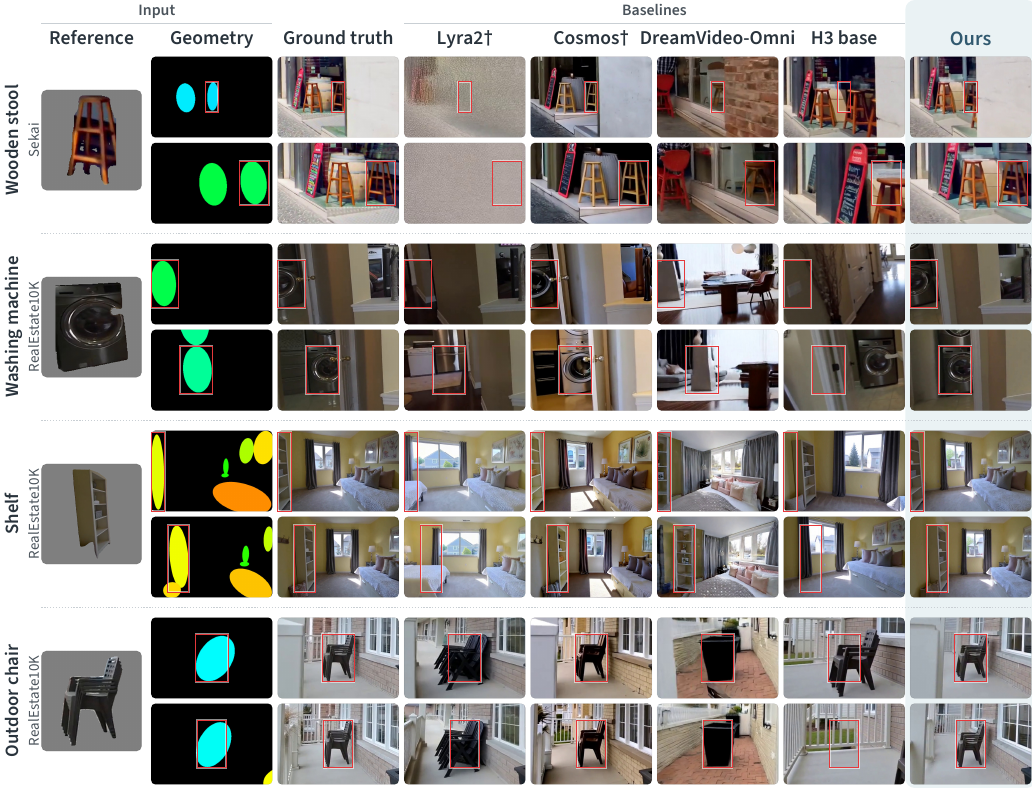}
\vspace{-0.4em}
\caption{\textbf{Additional baseline comparisons on four real-video clips.} Following the main-paper format, each row shows a zoom centered on the object's ground-truth location (red box), as it enters the view and after entry, with an identically cropped geometry flow. $^\dagger$Methods without reference-image input; Cosmos additionally receives dense source depth. In the wooden stool clip, Lyra2 loses the street scene.}
\label{fig:baselines_supp}
\end{figure*}

Four supplementary figures extend the main-paper visual analysis: \cref{fig:budget} visualizes elastic latent readout (asset LoD) from a single checkpoint across the $100\%$, $50\%$, and $25\%$ token budgets; \cref{fig:world_examples,fig:worlds_more} present fourteen full-budget renderings across five held-out real-video clips and nine multi-asset authored 3D scenes under occlusion and large viewpoint shifts; and \cref{fig:baselines_supp} extends the real-video baseline comparison to four additional clips. Full-resolution video playback, synchronized Canvas comparisons, and additional multi-instance reference swaps across pottery, clocks, vases, and lamps are provided on \suppref{our \href{\projecturl}{project page} and demo video}{the accompanying offline supplementary webpage and demo video}.

\section{Limitations}
\label{sec:supp_limits}
\paragraph{Detail attenuation at reduced asset LoD.}
Downsampling cached asset latents via elastic latent readout trades fine surface texture and color saturation for shorter sequence lengths (Ref-DINO-Tracked moves from $0.759$ at $100\%$ to $0.731$ at $50\%$ and $0.736$ at $25\%$; \cref{fig:budget}), softening high-frequency details relative to a dense per-frame canvas.

\paragraph{Implicit camera control from sparse geometry flows.}
Without explicit camera parameters, global camera motion is inferred solely from the projected geometry flow; when proxies are sparse or camera motion is rapid, trajectory accuracy remains less precise than explicit camera-controlled or dense-depth conditioning.

\end{document}